# Consideration on Example 2 of "An Algorithm of General Fuzzy InferenceWith The Reductive Property"

**Son-Il KWAK[1 *], Oh-Chol GWON[2], Chung-Jin KWAK[3]**

[1] School of Information Science, Kim Il Sung University, Pyongyang, D.P.R of Korea, ryongnam18@yahoo.com
[2] Center of Natural Science Research, Kim Il Sung University, Pyongyang, D.P.R of Korea, ryongnnam22@yahoo.com
[3] School of Electricity and Automation, Kim Il Sung University, Pyongyang, D.P.R of Korea, ryongnnam13@yahoo.com

[*] Corresponding author

**A B S T R A C T**

---

**In this paper, we will show that (1) the results about the fuzzy reasoning algoritm obtained in the paper "Computer Sciences Vol. 34, No.4, pp.145-148, 2007" according to the paper "IEEE Transactions On systems, Man and cybernetics, 18, pp.1049-1056, 1988" are correct; (2) example 2 in the paper "An Algorithm of General Fuzzy Inference With The Reductive Property" presented by He Ying-Si, Quan Hai-Jin and Deng Hui-Wen according to the paper "An approximate analogical reasoning approach based on similarity measures" presented by Tursken I.B. and Zhong zhao is incorrect; (3) the mistakes in their paper are modified and then a calculation example of FMT is supplemented.**

---

## 1. Introduction

Zadeh proposed CRI principle[1], Wang tripleI principle (TIP) with total inference rules of fuzzy reasoning method, [2] and Yeung and Tsang [3] a similarity based fuzzy reasoning method, respectively. HE, QUAN and DENG proposed an algorithm of general fuzzy inference with the reductive property [4], and Chen proposed a fuzzy reasoning method of fuzzy system with many rules [5]. Tursken and Zhong proposed an approximate analogical reasoning approach based on similarity measures[6]. Yuan and Lee[7] showed that results about the triple I method for fuzzy reasoning obtained in Liu [8] are correct and the Example 2.1 in ref. 8 is incorrect. In this paper, we show that example 2 in He et al. [4] is incorrect; and the mistakes in their paper are modified and then a calculation example of FMT is supplemented.

Generally fuzzy reasoning is fuzzy modus ponens (FMP) in the fuzzy system with m inputs 1 output n rules.

General form of the fuzzy modus ponens in the paper [6] is as follows.

$$\text{Premise 1;}\quad \begin{array}{l} \text{if } x_{11}=A_{11} \text{ and } x_{12}=A_{12} \text{ and } \cdots\ x_{1m}=A_{1m} \text{ then } y_1=B_1 \\ \qquad\vdots \qquad\qquad\qquad \vdots \\ \text{if } x_{n1}=A_{n1} \text{ and } x_{n2}=A_{n2} \text{ and } \cdots\ x_{nm}=A_{nm} \text{ then } y_n=B_n \end{array} \tag{1}$$

$$\text{Premise 2:}\quad x_1=A_1^* \ \ and\ \ x_2=A_2^* \ \ and\ \ \cdots\ \ x_m=A_m^*$$

$$\text{Conclusion:}\quad y=B^*$$

General form of Fuzzy Modus Tollens (FMT) in the paper [6] is as follows.

$$\text{Premise 1;}\quad \begin{array}{l} \text{if } x_{11}=A_{11} \text{ and } x_{12}=A_{12} \text{ and } \cdots\ x_{1m}=A_{1m} \text{ then } y_1=B_1 \\ \quad\vdots \qquad\qquad \vdots \\ \text{if } x_{n1}=A_{n1} \text{ and } x_{n2}=A_{n2} \text{ and } \cdots\ x_{nm}=A_{nm} \text{ then } y_n=B_n \end{array} \tag{2}$$

Premise 2; $y=B^*$

Conclusion: $x_1=A_1^*$ *and* $x_2=A_2^*$ *and* $\cdots$ $x_m=A_m^*$

where $A_{ij}$ and $A_j^*$ ( $j=1,\ 2,\ \dots,\ m$, $i=1,\ 2,\ \dots,\ n$) are fuzzy sets defined in the universe of discourse X, $B_i$ ( $i=1,\ 2,\ \dots,\ n$ ) and $B^*$ are fuzzy sets defined in the universe of discourse Y.

**Definition 1** [4,6]

Assume $S: F(X)\times F(X)\to[0,1]$.

If

$$S_1: \forall A\in F(X): S(A,A)=1,$$
$$S_2: \forall A\in P(X): S(A,A^c)=0$$
$$S_3: \forall A,B,C,D\in F(X)$$

and $\int_X |A(x)-B(x)|dx \ge \int_X |C(x)-D(x)|dx$, then $S(A,B)\le S(C,D)$ is satisfied.

Particularly for 3 fuzzy sets, if $A\subseteq B\subseteq C$, then $S(A,C)\le\min\{S(A,B),\ S(B,C)\}$ and S is called s imilarity defined on fuzzy set *F*(*X*).

In the fuzzy system with m inputs 1output n rules, definition for reductive property of fuzzy inference met hod is as follows;

**Definition 2** [2] Reductive property of FMP problem in fuzzy system with m inputs 1output n rules.

For fuzzy inference form (1) that has several premises, given $A_j^*=A_{ij}$ for $i=1,\ 2,\ \dots,\ n$, if $B^*=B_i\ \ (\forall j,\ j=1,\ 2,\ \dots,\ m)$ is satisfied, we call that this algorithm satisfies reductive property of FMP p roblem in the fuzzy system with m inputs 1output n rules.

**Definition 3** [2] Reductive property of FMT problem in the fuzzy system with m inputs 1output n rules.

For fuzzy inference form (2) that has several premises, given $B^*=B_i$ ( $i=1,\ 2,\ \dots,\ n$), if

$$A_j^*=A_{ij}\,(\forall j,\ j=1,\ 2,\ \dots,\ m) \tag{3}$$

is satisfied, then we call that this algorithm satisfies reductive property of FMT problem in the fuzzy system wit h m inputs 1output n rules.

## 2. Example 2 in the Paper[4] is incorrect.

In this section, we consider the fuzzy set in case that fuzzy set and input change on vertical axis not horizontal axis by using the example presented in the paper [4, 6].

Assume that the considered fuzzy set, 2 inputs 1 output 2 rules, is given as follows [2, 4].

$$\begin{array}{l} R_1: \text{if } x_1 \text{ is } A_{11}(x_1) \text{ and } x_2 \text{ is } A_{12}(x_2) \text{ then } y \text{ is } B_1(y) \\ R_2: \text{if } x_1 \text{ is } A_{21}(x_1) \text{ and } x_2 \text{ is } A_{22}(x_2) \text{ then } y \text{ is } B_2(y) \end{array} \tag{4}$$

where $X_i=Y=[0,\ 1],\ \ A_{ij}(x_i),\ \ A_j^*(x_j)\in F(X_i),\ \ B_i(y),\ \ B^*(y)\in F(Y)\quad i,\ j=1,2$ are the fuzzy set of antecedent of the rule, input, fuzzy set of consequent of the rule, and input respectively. For fuzzy rule 1 and 2 in formula (1), fuzzy set of the antecedent, fuzzy set of consequent and new input fuzzy set are as formula (5).

$$A_{11}(x_1) = \frac{1}{3}(3 - x_1), \quad A_{12}(x_2) = \frac{1}{3}(x_2 + 2), \quad B_1(y) = 1 - y$$
$$A_{21}(x_1) = 1 - x_1, \quad A_{22}(x_2) = 1 - \frac{1}{2}x_2, \quad B_2(y) = \frac{1}{2}(1 + y) \tag{5}$$
$$A_1^*(x_1) = 1 - x_1, \quad A_2^*(x_2) = 1 - \frac{1}{2}x_2, \quad B^*(y) = 1 - \frac{1}{2}(1 + y)$$

Consider the problem obtaining the reasoning result $B^*$ for example 2 in the paper [4]. We use similarity SM (A, B) in the paper [6].

$$SM(A, B) = \left(1 + \frac{1}{b-a}\int_a^b |A(u) - B(u)|du\right)^{-1} \tag{6}$$

In formula (6) $u \in U$ is continuous domain of definition and we calculate the sub-reasoning result in two types ${}_{type\,1}B_i^*$ and ${}_{type\,2}B_i^*$.

$${}_{type1}B_i^* = SM(A_{i1},\ A_1^*)B_i \cap SM(A_{i2},\ A_2^*)B_i \tag{7}$$

$${}_{type\,2}B_i^* = \min\left\{1,\ \frac{B_i}{SM(A_{i1},\ A_1^*)}\right\} \cap \min\left\{1,\ \frac{B_i}{SM(A_{i2},\ A_2^*)}\right\} \tag{8}$$

Using formula (7) and (8) sub-reasoning results are as follows.

$$\begin{cases} {}_{type\,1}B_1^* = SM(A_{11},\ A_1^*)B_1 \cap SM(A_{12},\ A_2^*)B_1 = \frac{3}{4}(1-y) \\ {}_{type\,1}B_2^* = SM(A_{21},\ A_1^*)B_2 \cap SM(A_{22},\ A_2^*)B_2 = \frac{1}{2}(1+y) \end{cases} \tag{9}$$

$$\begin{cases} {}_{type\,2}B_1^* = \min\left\{1, \frac{B_1}{SM(A_{11},\ A_1^*)}\right\} \cap \min\left\{1,\ \frac{B_1}{SM(A_{12},\ A_2^*)}\right\} = \begin{cases} 1, & y < \frac{1}{13} \\ \frac{13}{12}(1-y), & y \geq \frac{1}{13} \end{cases} \\ {}_{type\,2}B_2^* = \min\left\{1, \frac{B_2}{SM(A_{21},\ A_1^*)}\right\} \cap \min\left\{1,\ \frac{B_2}{SM(A_{22},\ A_2^*)}\right\} = \frac{1}{2}(y+1) \end{cases} \tag{10}$$

According to the paper [4, 6], the calculation results of formula (9) and (10) are correct. The fuzzy reasoning results of the paper [4, 6] are expressed as ${}_{type\ 1}B^*$ (11) and ${}_{type\,2}B^*$ (12) respectively, but the formula (11) and (12) are incorrect respectively.

$${}_{type\ 1}B^* = {}_{type\ 1}B_1^* \cup {}_{type\ 1}B_2^* = \begin{cases} \frac{3}{4}(1-y), & y \leq \frac{1}{5} \\ \frac{1}{2}(y+1), & y > \frac{1}{5} \end{cases} \tag{11}$$

$$
{}_{type\ 2}B^* = {}_{type\ 2}B_1^* \cup {}_{type\ 2}B_2^* = \begin{cases} 1, & y < \frac{1}{13} \\ \frac{13}{12}(1-y), & \frac{1}{13} \le y < \frac{7}{19} \\ \frac{1}{2}(y+1), & y \ge \frac{7}{19} \end{cases} \tag{12}
$$

The results of the formula (9) and (11) based on the paper [4, 6] are wrong. The recalculated result of formula (9) and (11) based on the paper [4, 6] is the formula (13) and (14) respectively in this paper.

$$
\begin{cases} {}_{type\ 1}B_{1new}^* = SM(A_{11}, A_1^*)B_1 \cap SM(A_{12}, A_2^*)B_1 = \frac{3}{4}(1-y) \\ {}_{type\ 1}B_{2new}^* = SM(A_{21}, A_1^*)B_2 \cap SM(A_{22}, A_2^*)B_2 = \frac{60}{73}(1+y) \end{cases} \tag{13}
$$

$$
{}_{type\ 1}B_{new}^* = {}_{type\ 1}B_{1\ new}^* \cup {}_{type\ 1}B_{2\ new}^* = \begin{cases} \frac{3}{4}(1-y), & y \le \frac{1}{3} \\ \frac{3}{8}(y+1), & y > \frac{1}{3} \end{cases} \tag{14}
$$

The results of the formula (10) and (12) based on the paper [4, 6] are wrong. The recalculated fuzzy reasoning results of formula (10) and (12) based on the paper [4, 6] are the same as formula (15) and (16) respectively in this paper.

$$
\begin{cases} {}_{type\ 2}B_{1new}^* = \min\left\{1, \frac{B1}{SM(A_{11}, A_1^*)}\right\} \cap \min\left\{1, \frac{B1}{SM(A_{12}, A_2^*)}\right\} = \begin{cases} 1, & y < \frac{13}{73} \\ \frac{73}{60}(1-y), & y \ge \frac{13}{73} \end{cases} \\ {}_{type\ 2}B_{2\ new}^* = \min\left\{1, \frac{B1}{SM(A_{21}, A_1^*)}\right\} \cap \min\left\{1, \frac{B2}{SM(A_{22}, A_2^*)}\right\} = \begin{cases} \frac{73}{120}(1+y), & y < \frac{47}{73} \\ 1, & y \ge \frac{47}{73} \end{cases} \end{cases} \tag{15}
$$

$$
{}_{type\ 2}B_{new}^* = {}_{type\ 2}B_{1\ new}^* \cup {}_{type\ 2}B_{2\ new}^* = \begin{cases} 1, & y < \frac{13}{73} \\ \frac{73}{60}(1-y), & \frac{13}{73} \le y < \frac{1}{3} \\ \frac{73}{120}(y+1), & \frac{1}{3} \le y < \frac{47}{73} \\ 1, & y \ge \frac{47}{73} \end{cases} \tag{16}
$$

From the above example, the fuzzy modus ponens (FMP) by similarity in the paper [4,6] also doesn't satisfy the reductive property in m inputs 1 output n rules. It is the same in fuzzy modus tollens.In the next section we consider FMT of the paper [4].

## 3. Calculation of FMT based on similarity of the paper [4]

Input of FMT according to formula (5) is as follows.

$$
B^*(y) = 1 - B_2(y) = \frac{1}{2}(1-y) \tag{17}
$$

The subreasoning calculation result is as the formula (18).

$$\begin{cases} {}_{type\,1}A^{*}_{11new} = SM(B_1,\ B^{*})A_{11} = \frac{4}{15}(3-x_1) \\ {}_{type\,1}A^{*}_{12new} = SM(B_1,\ B^{*})A_{12} = \frac{4}{15}(x_2+2) \\ {}_{type\,1}A^{*}_{21new} = SM(B_2,\ B^{*})A_{21} = \frac{2}{3}(1-x_1) \\ {}_{type\,1}A^{*}_{22new} = SM(B_2,\ B^{*})A_{22} = \frac{1}{3}(2-x_2) \end{cases} \tag{18}$$

The FMT calculation result of *type 1* is as the formula (19).

$$\begin{aligned} {}_{type\,1}A^{*}_{1new} &= {}_{type\,1}A^{*}_{11\,new} \cup {}_{type\,1}A^{*}_{21\,new} = \frac{4}{15}(3-x_1) \\ {}_{type\,1}A^{*}_{2new} &= {}_{type\,1}A^{*}_{12\,new} \cup {}_{type\,1}A^{*}_{22new} = \begin{cases} \frac{1}{3}(2-x_2), & x_2 \le \frac{2}{9} \\ \frac{4}{15}(x_2+2), & y > \frac{2}{9} \end{cases} \end{aligned} \tag{19}$$

Therefore, since ${}_{type\,1}A^{*}_{1new} \ne \overline{A_{21}}$, ${}_{type\,1}A^{*}_{2new} \ne \overline{A_{22}}$, FMT of *type 1* in the paper [4, 6] doesn't satisfied reductive property. The subreasoning calculation result of *type 2* is as the formula (20).

$$\begin{cases} {}_{type\;2}A^{*}_{11\,new} = \min\left\{1,\ \frac{A_{11}}{SM(B_1,\ B^{*})}\right\} = \begin{cases} 1, & x_1 < \frac{3}{5} \\ \frac{5}{12}(3-x_1), & x_1 \ge \frac{3}{5} \end{cases} \\ {}_{type\;2}A^{*}_{12\,new} = \min\left\{1,\ \frac{A_{12}}{SM(B_1,\ B^{*})}\right\} = \begin{cases} \frac{5}{12}(x_2+2), & x_2 < \frac{2}{5} \\ 1, & x_2 \ge \frac{2}{5} \end{cases} \\ {}_{type\;2}A^{*}_{21\,new} = \min\left\{1,\ \frac{A_{21}}{SM(B_2,\ B^{*})}\right\} = \begin{cases} 1, & x_1 < \frac{1}{3} \\ \frac{3}{2}(1-x_1), & x_1 \ge \frac{1}{3} \end{cases} \\ {}_{type\;2}A^{*}_{22\,new} = \min\left\{1,\ \frac{A_{21}}{SM(B_2,\ B^{*})}\right\} = \begin{cases} 1, & x_2 < \frac{2}{3} \\ \frac{3}{4}(2-x_2), & x_2 \ge \frac{2}{3} \end{cases} \end{cases} \tag{20}$$

The FMT calculation result of the *type 2* is as the formula (21).

$$\begin{aligned} {}_{type\;2}A^{*}_{1new} &= {}_{type\;2}A^{*}_{11\,new} \cup {}_{type\;2}A^{*}_{21\,new} = \begin{cases} 1, & x_1 \le \frac{3}{5} \\ \frac{5}{12}(3-x_1), & x_1 > \frac{3}{5} \end{cases} \\ {}_{type\;2}A^{*}_{2new} &= {}_{type\;2}A^{*}_{12\,new} \cup {}_{type\;2}A^{*}_{22new} = 1 \end{aligned} \tag{21}$$

Therefore, since ${}_{type\,2}A^{*}_{1new} \ne \overline{A_{21}}$, ${}_{type\,2}A^{*}_{2new} \ne \overline{A_{22}}$, FMT of *type 2* in the paper [4, 6] does not satisfied reductive property. From the above calculation, we can see that the fuzzy modus tollens (FMT) by similarity in the paper [4, 6] also doesn't satisfy the reductive property in m inputs 1 output n rules.

## 4. Conclusions

In this paper, we have shown that

(1) The results about the Algorithm of General Fuzzy Inference with The Reductive Property obtained in pa per [4] are correct.
(2) The Example 2 in paper [4] “An Algorithm of General Fuzzy Inference with The Reductive Property” is incorrect, also, the example in the paper [6] “An approximate analogical reasoning approach based on simi larity measures”.
(3) The mistakes in the paper [4] are modified and then the calculation of FMT is supplemented.